\documentclass[11pt,a4paper]{article}
\usepackage[hyperref]{acl2019}

\usepackage{times}
\usepackage{latexsym}
\usepackage{microtype}
\usepackage{graphicx}
\usepackage{subfigure}
\usepackage{booktabs} 
\usepackage{multirow}
\usepackage{url}
\usepackage{amsmath}
\usepackage{float}
\usepackage{booktabs}
\usepackage{amssymb}
\usepackage{mathtools}
\usepackage{capt-of}
\usepackage{cleveref}
\usepackage{algorithm}
\usepackage{algorithmic}
\usepackage{url}

\aclfinalcopy 
\def\aclpaperid{160} 

\title{Lifelong and Interactive Learning of Factual Knowledge in Dialogues}

\author{Sahisnu Mazumder,~ Bing Liu,~ Shuai Wang,~ Nianzu Ma \\
  Department of Computer Science, University of Illinois at Chicago, USA \\
  \texttt{sahisnumazumder@gmail.com,liub@uic.edu}\\
  \texttt{shuaiwanghk@gmail.com,jingyima005@gmail.com} \\}

\date{}

\begin{document}
\maketitle
\begin{abstract}
Dialogue systems are increasingly using \textit{knowledge bases} (KBs) storing real-world facts to help generate quality responses. However, as the KBs are inherently incomplete and remain fixed during conversation, it limits dialogue systems' ability to answer questions and to handle questions involving entities or relations that are not in the KB. In this paper, we make an attempt to propose an engine for \textit{Continuous and Interactive Learning of Knowledge} (CILK) for dialogue systems to give them the ability to continuously and interactively learn and infer new knowledge during conversations. With more knowledge accumulated over time, they will be able to learn better and answer more questions. Our empirical evaluation shows that CILK is promising. 
\end{abstract}

\vspace{-0.1cm}

\section{Introduction}
Dialogue systems, including  question-answering (QA) systems are now commonly used in practice. Early such systems were built mainly based on rules and information retrieval techniques \cite{banchs2012iris,ameixa2014luke,lowe2015ubuntu,serban2015survey}. 
Recent deep learning models \cite{vinyals2015neural,xing2017topic,li2017adversarial} learn from large corpora. However, since they do not use explicit knowledge bases (KBs), they often suffer from generic and dull responses \cite{xing2017topic,Young2018Augmenting}. KBs have been used to deal with the problem \cite{ghazvininejad2017knowledge,le2016lstm,Young2018Augmenting,long2017knowledge,zhou2018commonsense}. Many task-oriented dialogue systems \cite{eric2017key,madotto2018mem2seq} also use KBs to support information-seeking conversations. 

One major shortcoming of existing systems that use KBs is that the KBs are fixed once the dialogue systems are deployed. However, it is almost impossible for the initial KBs to contain all possible knowledge that the user may ask, not to mention that new knowledge appears constantly. It is thus highly desirable for dialogue systems to learn by themselves while in use, i.e., \textit{learning on the job} in \textit{lifelong learning}~\cite{ChenAndLiubook2018}. Clearly, the system can \textbf{(1)} extract more knowledge from the Web or other sources, and \textbf{(2)} learn directly from users during conversations. This paper focuses on the latter and makes an attempt to propose an engine for \textit{Continuous and Interactive Learning of Knowledge} (CILK) to give the dialogue system the ability to acquire/learn new knowledge from the user during conversation. Specifically, it focuses on learning new knowledge interactively from the user when the system is unable to answer a user's WH-question. The acquired new knowledge makes the system better able to answer future user questions, and no longer be limited by the fixed knowledge provided by the human developers.

The type of knowledge that the CILK engine focuses on is the facts that can be expressed as triples, ($h$, $r$, $t$), which means that the \textit{head entity} $h$ and the \textit{tail entity} $t$ can be linked by the \textit{relation} $r$. An example of a fact is (\textit{Boston}, \textit{LocatedInCountry}, \textit{USA}), meaning that \textit{Boston is located in USA}. This paper only develops the core engine. It does not study other dialogue functions like response generation, semantic parsing, fact extraction from user utterances, entity linking, etc., which have been studied extensively before and are assumed to be available for use. Thus, this paper works only with structured queries ($h$, $r$, $?$), e.g., (\textit{Boston}, \textit{LocatedInCountry}, ?) meaning ``\textit{In what Country is Boston located ?},'' or ($?$, $r$, $t$), e.g., (?, \textit{PresidentOf}, \textit{USA}) meaning ``\textit{Who is the President of USA?}'' It assumes that a semantic parser is available that can convert natural language queries from users into query triples. Similarly, it assumes an information extraction tool like OpenIE~\cite{angeli2015leveraging} is employed to extract facts as triples ($h$, $r$, $t$) from user's utterances during conversation.
Building a full-fledged dialogue system that can also learn during conversation is a huge undertaking and is out of the scope of this paper. We thus only investigate the core knowledge learning engine here. We also assume that the user has good intentions (i.e., user answers questions with 100\% conformity about the veracity of his/her facts)\footnote{We envision that the proposed engine is incorporated into a dialogue system in a multi-user environment. The system can perform cross-verification with other users by asking them whether the knowledge (facts) from a user is correct.}; but is not omniscient (opposed to the teacher-student learning setup).

\vspace{1mm}
\textbf{Problem Definition:} Given a user query / question ($h$, $r$, ?) [or (?, $r$, $t$)], where $r$ and $h$ (or $t$) may not be in the KB (i.e., unknown), our goal is two-fold: (i) \textit{answering the user query} or \textit{rejecting the query to remain unanswered} in the case when the correct answer is believed to not exist in the KB and (ii) \textit{learning / acquiring some knowledge (facts) from the user to help the answering task}. We only focus on the setting where the query cannot be answered \textit{directly} with the current KB and need inference over existing facts, as considering structured query, it's trivial to retrieve the answer if the answer triple is already in KB. We further distinguish two types of queries: \textbf{(1)} \textit{closed-world queries}, where $h$ (or $t$) and $r$ are known to the KB, and \textbf{(2)} \textit{open-world queries}, where either one or both $h$ (or $t$) and $r$ are unknown to the KB. 

It is easy to see that the problem is essentially a \textit{lifelong learning} problem~\cite{ChenAndLiubook2018}, where each query to be processed is a task and the knowledge gained is retained in the KB. To process a new query/task, the knowledge learned and accumulated from the past queries can be leveraged. 

For each new open-world query, the proposed approach works in two steps: 

\vspace{0.5mm}
\textbf{Step 1 - Interact with the user:} It converts open-world queries (2) to closed-world queries (1) by asking the user questions related to $h$ (or $t$) and $r$ to make them known to the KB (added to KB). The reason for the conversion will be clear below. The user answers, called \textit{supporting facts} (SFs), are the new knowledge to be added to  KB. This step is also called \textit{interactive knowledge learning}. Note, closed-world queries (1) do not need this step. 

\vspace{1mm}
\textbf{Step 2 - Infer the query answer}: It solves closed-world queries (1) by inferring the query answer. The main idea is to use each entity $e$ in the KB to form a candidate triple ($h$, $r$, $e$) (or ($e$, $r$, $t$)), which is then scored. The entity $e$ with the highest score is predicted as the answer of the query. 

Scoring each candidate is modeled as a \textit{knowledge base completion} (KBC) problem~\cite{lao2010relational,bordes2011learning}. KBC aims to infer new facts (knowledge) from existing facts in a KB and is defined as a \textit{link prediction} problem: Given a query triple, ($e$, $r$, ?) [or (?, $r$, $e$)], it predicts a tail entity $t_{true}$ [head entity $h_{true}$] which makes the query triple true and thus should be added to the KB. KBC makes the \textit{closed-world} assumption that $h$, $r$ and $t$ are all \textit{known} to exist in the KB \cite{lao2011random,bordes2011learning,bordes2013translating,nickel2015review}. This is not suitable for knowledge learning in conversations because in a conversation, the user can ask or say anything, which may contain entities and relations that are not in the KB. CILK removes the closed-world assumption and allows all $h$ (or $t$) and/or $r$ to be \textit{unknown} (not in the KB). Step 1 above basically asks the user questions to make $h$ (or $t$) and/or $r$ known to the KB. Then, an existing KBC model as a query inference model can be applied to retrieve an answer entity from KB.

\begin{table}[t]
	\label{my-label}
	\small
	\begin{tabular}{lp{5.57cm}}
		\textbf{\small USER:} &  \textcolor{green!25!black}{(\textit{Boston}, \textit{LocatedInCountry}, \textit{?})} ``\textcolor{blue!55!black}{In what Country is Boston located?}'' ~~~~\hfill[\textcolor{red}{\textbf{Query}}] \\
		
		\textbf{\small CILK:} & \textcolor{red!45!black}{I do not know what ``\textbf{located in Country}'' means? Can you provide me an example?} \\
		& \hfill[\textit{\textbf{Ask for Clue}}] \\
		
		\textbf{\small USER:} & \textcolor{green!25!black}{(\textit{London}, \textit{LocatedInCountry}, \textit{UK})}. ``\textcolor{blue!55!black}{London is located in UK.}''
		\hfill[\textcolor{green!67!black}{\textbf{SF1}}]\\
		
		\textbf{\small CILK:} & \textcolor{red!45!black}{Got it. Can you tell me a fact about ``\textbf{Boston}''?} \hfill[\textit{\textbf{Ask for Entity Fact}}] \\
		
		\textbf{\small USER:} & \textcolor{green!25!black}{(\textit{Harvard University}, \textit{UniversityLocatedIn}, \textit{Boston})}. ``\textcolor{blue!55!black}{Harvard university is located in Boston.}''~~~~~~~~~~\hfill[\textcolor{green!67!black}{\textbf{SF2}}]\\
		\textbf{\small CILK:} & \textcolor{green!25!black}{(\textit{Boston}, \textit{LocatedInCountry}, \textit{USA})} ``\textcolor{red!45!black}{Boston is located in USA.}" \hfill[\textcolor{blue}{\textbf{Answer}}]\\
	\end{tabular}
	\vspace{-0.1cm}
	\captionof{figure}{\small An example of interactive learning and inference. Note that CILK only works with triples. Each triple above is assumed to be extracted from the sentence after it. \textit{\textbf{Ask for Clue}} and  \textit{\textbf{Ask for Entity Fact}} are interaction query types, discussed in Sec.~3. SF denotes supporting fact.}
	\normalsize
	\vspace{-0.3cm}
\end{table}

Figure 1 shows an example. CILK acquires supporting facts SF1 and SF2 to accomplish the goal of \textit{knowledge learning} and utilizes these pieces of knowledge along with existing KB facts to answer the user query (i.e., to infer over the query relation "\textit{LocatedInCountry}"). CILK aims to achieve these two sub-goals. The new knowledge (SFs) is added to the KB for future use\footnote{The inferred query answer is not added to the KB as it may be incorrect. But it can be added in a multi-user environment through cross-verification (see footnote 1 and Sec. 4).}. We evaluate CILK using two real-world KBs: \textit{Nell} and \textit{WordNet} and obtain promising results. 

\section{Related Work}
\label{sec:RW}
To the best of our knowledge, no existing system can perform the proposed task. We reported a priliminary research in~\cite{mazumder2018towards}. 

CILK is related to interactive language learning \cite{wang2016learning,wang2017naturalizing}, which is mainly about language grounding, not about knowledge learning. 
\citet{li2016dialogue,li2016learning} and \citet{zhang2017listen} train chatbots using human teachers who can ask and answer the chatbot questions.  
\citet{Otsuka2017Lexical},~\citet{Otsuka2013Generating},~\citet{Ono2016Toward} and~\citet{Komatani2018Question} allow a system to ask the user whether its prediction of category of a term is correct or not. Compared to these works, CILK performs interactive knowledge learning and inference (over existing and acquired knowledge) while conversing with users after the dialogue system has been deployed (i.e., \textit{learning on the job}~\cite{ChenAndLiubook2018}) without any teacher supervision or help. 

NELL \cite{mitchell2015never} updates its KB using facts extracted from the Web (complementary to our work). We do not do Web fact extraction.  

KB completion (KBC) has been studied in recent years~\cite{lao2011random,bordes2011learning,bordes2015large,mazumder2017context}. But they mainly handle facts with known entities and relations. \citet{neelakantan2015compositional} work on fixed unknown relations with known embeddings, but does not allow unknown entities. \citet{xiong2018one} also deal with queries involving unknown relations, but known entities in the KB. \citet{shi2017open} handles unknown entities by exploiting an external text corpus. None of the KBC methods perform conversational knowledge learning like CILK.


\section{Proposed Technique}

As discussed in Sec.~1, given a query ($e$, $r$, ?) [or ($?$, $r$, $e$)]\footnote{Either $e$ or $r$ or both may not exist in the KB} from the user, CILK interacts with the user to acquire supporting facts to answer the query. Such an interactive knowledge learning and inference task is realized by the cooperation of three primary components of CILK:~  \textbf{\textit{Knowledge base}} (KB) $\mathcal{K}$, \textbf{\textit{Interaction Module}} $\mathcal{I}$ and \textbf{\textit{Inference Model}} $\mathcal{M}$. The interaction module $\mathcal{I}$ decides whether to ask or not and formulates questions to ask the user for supporting facts. The acquired supporting facts are added to the KB $\mathcal{K}$ and used in training the Inference Model $\mathcal{M}$ which then performs inference over the query (i.e., answers the query).

In the following subsections, we formalize the interactive knowledge learning problem (Sec. 3.1), describe the Inference Model $\mathcal{M}$ (Sec. 3.2) and discuss how CILK interacts and processes a query from the user (Sec. 3.3). 

\vspace{-0.1cm}
\subsection{Problem Formulation}
CILK's KB $\mathcal{K}$ is a triple store \{($h$, $r$, $t$)\} $\subseteq \mathcal{E} \times \mathcal{R}  \times \mathcal{E}$, where $\mathcal{E}$ is the entity set and $\mathcal{R}$ is the relation set. Let $q$ be a query of the form ($e$, $r$, ?) [or ($?$, $r$, $e$)] issued to CILK, where $e$ is termed as \textit{query entity} and $r$ 
as the \textit{query relation}.  If $e \notin \mathcal{E}$ and/or $r \notin \mathcal{R}$ (we also say $e, r \notin \mathcal{K}$), we call $q$ an \textit{open-world} query. Otherwise, $q$ is referred to as a \textit{closed-world} query, i.e., both $e$ and $r$ exist in $\mathcal{K}$.
Given $\mathcal{K}$ and a query $q$, the query inference task is defined as follows: If $q$ is of the form ($e$, $r$, ?), the goal is to predict a tail entity $t_{true} \in \mathcal{E}$ such that ($e$, $r$, $t_{true}$) holds. We call such $q$ a \textit{tail query}. If $q$ is of the form (?, $r$, $e$), the goal is to predict a head entity $h_{true} \in \mathcal{E}$ such that ($h_{true}$, $r$, $e$) holds. We call such $q$ a \textit{head query}.  
In the open-world setting, it's quite possible that the answer entity $t_{true}$ (for a tail query) or $h_{true}$ (for a head query) does not exist in the KB (in $\mathcal{E}$). In such cases, the inference model $\mathcal{M}$ cannot find the true answer. We thus further extend the goal of query inference task to either finding answer entity $t_{true}$ ($h_{true}$) for $q$ or rejecting $q$ to indicate that the answer does not exist in $\mathcal{E}$.

Given an open-world (head / tail) query $q$ from user $u$, CILK interacts with $u$ to acquire a set of supporting facts (SFs) [i.e., a set of \textit{clue} triples $C_r$ involving query relation $r$ and/or a set of \textit{entity fact} triples $F_e$ involving query entity $e$] for learning $r$ and $e$ (discussed in Sec 3.3). In Figure 1, (\textit{London}, \textit{LocatedInCountry}, \textit{UK}) is a clue of query relation ``\textit{LocatedInCountry}" and (\textit{Harvard  University}, \textit{UniversityLocatedIn}, \textit{Boston}) is an entity fact involving query entity ``\textit{Boston}". In this interaction process, CILK decides and asks questions to the user for knowledge acquisition in multiple dialogue turns (see Figure 1). This is \textbf{step 1} as discussed in Sec.~1 and will be further discussed in Sec.~3.3. 

Once SFs are gathered, it uses $(\mathcal{K} \cup C_r \cup F_e)$ to infer $q$, which is \textbf{step 2} in Sec.~1 and will be detailed in Sec.~3.2. We refer to the whole interaction process involving multi-turn knowledge acquisition followed by the query inference step as a \textit{dialogue session}. In summary, CILK is assumed to operate in multiple dialogue sessions with different users and acquire knowledge 
in each session and thereby, continuously learns new knowledge over time. 

\subsection{Inference Model}
Given a query $q$, the Inference Model $\mathcal{M}$ attempts to infer $q$ by predicting the answer entity from $\mathcal{E}$. In particular, it selects each entity $e_i \in \mathcal{E}$ and forms $|\mathcal{E}|$ number of candidate triples \{$d_1$, ..., $d_{|\mathcal{E}|}$\}, where $d_i$ is of the form ($e$, $r$, $e_i$) for a tail query [or ($e_i$, $r$, $e$) for a head query] and then score each $d_i$ to quantify the relevancy of $e_i$ of being an answer to $q$. The top ranked entity $e_i$ is returned as the predicted answer of $q$. We deal with the case of query rejection by $\mathcal{M}$ later.

We use the neural knowledge base embedding (KBE) approach \cite{bordes2011learning,bordes2013translating,yang2014embedding} to design $\mathcal{M}$. Given a KB represented as a triple store, a neural KBE method learns to encode relational information in the KB using low-dimensional representations (embeddings) of entities and relations and uses the learned representations to predict the correctness of unseen triples. In particular, the goal is to learn representations for entities and relations such that valid triples receive high scores (or low energies) and invalid triples receive low scores (or high energies) defined by a scoring function $S(.)$. The embeddings can be learned via a neural network. In a typical (linear) KBE model, given a triple ($h$, $r$, $t$), input entity $h$, $t$ and relation $r$ correspond to high-dimensional vectors (either ``one-hot" index vector or ``n-hot" feature vector) $\mathbf{x}_h$, $\mathbf{x}_t$ and $\mathbf{x}_r$ respectively, which are then projected into low dimensional vectors $\mathbf{v}_h$, $\mathbf{v}_t$ and $\mathbf{v}_r$ using an entity embedding matrix $W_E$ and relation embedding matrix $W_R$ as given by- $\mathbf{v}_h = W_E~\mathbf{x}_h$, $\mathbf{v}_r = W_R~\mathbf{x}_r$ and $\mathbf{v}_t = W_E~\mathbf{x}_t$. The scoring function $S(.)$ is then used to compute a validity score $S(h, r, t)$ of the triple.

Any KBE model can be used for learning $\mathcal{M}$. For evaluation, we adopt DistMult \cite{yang2014embedding} for its state-of-the art performance over many other KBE models \cite{kadlec2017knowledge}. The scoring function of DistMult is defined as follows:
\vspace{-0.2cm}
\begin{equation}\label{key}
\small
S(h, r, t) = \mathbf{v}_h^T diag(\mathbf{v}_r) \mathbf{v}_t = \sum_{i=1}^N \mathbf{v_h}[i] \mathbf{v}_r[i] \mathbf{v}_t[i]
\vspace{-0.2cm}
\end{equation}
where $diag(\mathbf{v}_r)$ is the diagonal matrix in $\mathbf{v}_r$. 

The parameters of $\mathcal{M}$, i.e., $W_E$ and $W_R$, are learned by minimizing a margin-based ranking objective $\mathcal{L}$, which encourages the scores of positive triples to be higher than those of negative triples:
\vspace{-0.15cm}
\begin{equation}\label{key}
\small
\begin{split}
\mathcal{L} = \sum_{d \in D^+} \sum_{d' \in D^{-}} max\{S(d') - S(d) + 1, 0\}
\vspace{-0.3cm}
\end{split}
\end{equation}
where, $D^+$ is a set of triples observed in $\mathcal{K}$, treated as positive triples.  $D^{-}$ is a set of negative triples obtained by corrupting either head entity or tail entity of each +ve triple ($h$, $r$, $t$) in $D^{+}$ by replacing it with a randomly chosen entity $h'$ and $t'$ respectively from $\mathcal{K}$ such that the corrupted triples ($h'$, $r$, $t$), ($h$, $r$, $t'$) $\notin \mathcal{K}$. Note, $\mathcal{M}$ is trained continuously by sampling a set of +ve triples and correspondingly constructing a set of -ve triples as the KB expands with acquired supporting facts to improve its inference capability over new queries (involving new query relations and entities). Thus, the embedding matrices $W_E$ and $W_R$ also grow linearly over time.

\vspace{2mm}
\textbf{Rejection in KB Inference.} For a query with no answer entity existing in $\mathcal{K}$, CILK attempts to reject the query from being answered. To decide whether to reject the query or not, CILK maintains a \textbf{threshold buffer} $\mathcal{T}$ that stores entity and relation specific prediction thresholds and updates it continuously over time, as described below.

Besides the dataset for training $\mathcal{M}$, CILK also creates a validation dataset $D_{vd}$, consisting of a set of validation query tuples of the form ($q$, $E^+$, $E^-$). Here, $q$ is either a head or tail query involving query entity $e$ and relation $r$, $E^+ =$\{$e_1^+, .., e_p^+$\} is the set of $p$ positive (true answer) entities in $\mathcal{K}$ and $E^- =$\{$e_1^-, .., e_n^-$\} is the set of $n$ negative entities randomly sampled from $\mathcal{K}$ such that $E^+ \cap E^- = \emptyset$. 

Let $D_{vd}^e = \{(q, E^+, E^-)~|~(q, E^+, E^-) \in D_{vd},~ e\in q\}$ be the validation query tuple set involving entity $e$ and $D_{vd}^r = \{(q, E^+, E^-)~|~(q, E^+, E^-) \in D_{vd},~ r\in q \}$ be the validation query tuple set involving relation $r$.  Then, we compute $\mathcal{T}[z]$, (i.e., prediction threshold for $z$, where $z$ is either $e$ or $r$) as the average of the mean scores of triples involving +ve entities and mean scores of triples involving -ve entities, computed over all $q$ in $D_{vd}^z$, given by-
\vspace{-0.2cm}
\begin{equation}\label{key}
\small
\begin{split}
\mathcal{T}[z] = \frac{1}{2|D_{vd}^z|} \sum_{(q, E^+, E^-)\in D_{vd}^z} \mu_E^+ + \mu_E^-
\end{split}
\vspace{-0.3cm}
\end{equation}
where $\mu_E^+ = \frac{1}{|E^+|} \sum_{e_i^+ \in E^+} S(q, e_i^+)$ and $\mu_E^- = \frac{1}{|E^-|} \sum_{e_i^- \in E^-} S(q, e_i^-)$. Here, $S(q, e_i^+) = S(e, r, e_i^+)$ if $q$ is a tail query and $S(e_i^+, r, e)$ if $q$ is a head query. $S(q, e_i^-)$ can be explained in a similar way.

Given a head or tail query $q$ involving query entity $e$ and relation $r$, we compute the prediction threshold $\mu_q$ for $q$ as $\mu_q= max\{\mathcal{T}[e], \mathcal{T}[r], 0\}$. 

\vspace{2mm}
\textbf{Inference Decision Making.} If $\widetilde{e} \in \mathcal{E}$ is the predicted answer entity by $\mathcal{M}$ for query $q$ and $S(q, \widetilde{e}) > \mu_q$, CILK responds to user with answer  $\widetilde{e}$. Otherwise, $q$ gets rejected.

\setlength{\textfloatsep}{0.1cm}
\setlength{\floatsep}{0.1cm}
\begin{algorithm}[tb]
	\small
	\caption{\small CILK Knowledge Learning and Inference}
	\label{alg:example}
	\begin{flushleft}
		\textbf{Input:} query $q_j=(e, r, ?)~or~(?, r, e)$ issued by user at session-$j$;
		$\mathcal{K}_{j}$: CILK's KB at session-$j$;
		$\mathcal{P}_{j}$: Performance Buffer at session-$j$;
		$\mathcal{T}_{j}$: Threshold Buffer at session-$j$;
		$\mathcal{M}_{j}$: trained Inference Model at session-$j$; 
		$\alpha$: probability of treating an acquired supporting fact as training triple; $\rho$: \% of entities or relations in $\mathcal{K}_{j}$ that belong to the diffident set. \\
		\textbf{Output:} $\widetilde{e}:$ predicted entity as answer of query $q_j$ in session-$j$. 
	\end{flushleft}
	\vspace{0.1cm}
	\begin{algorithmic}[1]
		\IF{$r \notin \mathcal{K}_j$~ or~ \textbf{\texttt{IsDiffident}}($r$, $\mathcal{P}_j$, $\rho$)}  %
		\STATE $C_r~\leftarrow$ \textbf{\texttt{AskUserforCLUE}}($r$)  \COMMENT{acquire supporting facts to learn $r$'s embedding}
		\ENDIF
		\IF{$e \notin \mathcal{K}_j$~ or~ \textbf{\texttt{IsDiffident}}($e$, $\mathcal{P}_j$, $\rho$)}  %
		\STATE $F_e~\leftarrow$ \textbf{\texttt{AskUserforEntityFacts}}($e$)  \COMMENT{Acquire supporting facts to learn $e$'s embedding}
		\ENDIF
		
		\IF{$C_r \neq \emptyset$}  %
		\STATE $\mathcal{K}_{(j+\frac{1}{2})}\leftarrow$ Add clue triples from $C_r$ into $\mathcal{K}_j$ and randomly mark  $\alpha$\% of $C_r$ as training triples and (1-$\alpha$)\% as validation triples respectively in $\mathcal{K}_j$.
		\ENDIF 
		\IF{$F_e \neq \emptyset$}  %
		\STATE $\mathcal{K}_{j+1}\leftarrow$ Add fact triples from $F_e$ into $\mathcal{K}_{(j+\frac{1}{2})}$ and randomly mark  $\alpha$ \% of these triples as training triples and (1-$\alpha$) \% as validation triples.
		\ENDIF
		\STATE $D_{tr}^r$, $D_{vd}^r\leftarrow$ \textbf{\texttt{SampleTripleSet}}($\mathcal{K}_{j+1}$, $r$)
		\STATE $D_{tr}^e$, $D_{vd}^e\leftarrow$ \textbf{\texttt{SampleTripleSet}}($\mathcal{K}_{j+1}$, $e$)
		\STATE $\mathcal{M}_{j+1} \leftarrow$
		\textbf{\texttt{TrainInfModel}}($\mathcal{M}_j$, $D_{tr}^r\cup D_{tr}^e$)
		\STATE $\mathcal{P}_{j+1}, \mathcal{T}_{j+1} \leftarrow$
		\textbf{\texttt{UpdatePerfandThreshBuffer}}\\~~~~~~~~~~~~~~~~~~~~~~~~~~~~~($\mathcal{M}_{j+1}$,~ $(D_{vd}^r \cup D_{vd}^e)$,~ $\mathcal{P}_{j}$,~ $\mathcal{T}_{j}$) \\
		\STATE $\widetilde{e} \leftarrow$
		\textbf{\texttt{PredictAnswerEntity}}($\mathcal{M}_{j+1}$, $q_j$, $\mathcal{T}_{j+1}$)
	\end{algorithmic}
	\normalsize
\end{algorithm}
\setlength{\textfloatsep}{0.4cm}
\setlength{\floatsep}{0.4cm}

\subsection{Working of CILK}
Given a query $q$ involving unknown query entity $e$ and/or relation $r$, CILK has to ask the user to provide supporting facts to learn embeddings of $e$ and $r$ in order to infer $q$. However, the user in a given session can only provide very few supporting facts, which may not be sufficient for learning good embeddings of $e$ and $r$. Moreover, to accumulate a sufficiently good validation dataset for learning $\mathcal{T}[e]$ and $\mathcal{T}[r]$, CILK needs to gather more triples from users involving $e$ and $r$. But, asking for SFs for any entity and/or relation can be annoying to the user and also, is unnecessary if CILK has already learned good emmbeddings of that entity and/or relation (i.e., CILK has performed well in predicting true answer entity for queries involving that entity and/or relation in past dialogue sessions with other users). Thus, it is more reasonable to ask for SFs for the \textit{known} entities and/or relations \textit{for which CILK is not confident about} performing inference accurately, \textit{besides the unknown ones}.

To minimize the rate of user interaction and justify the knowledge acquisition process, CILK uses a \textbf{performance buffer} $\mathcal{P}$ to store the performance statistics of CILK in past dialogue sessions. We use Mean Reciprocal Rank (MRR)
to measure the performance of $\mathcal{M}$ (discussed in Sec. 4.1). In particular, $\mathcal{P}[e]$ and $\mathcal{P}[r]$ denote the avg. MRR achieved by $\mathcal{M}$ while answering queries involving $e$ and $r$ respectively, evaluated on the validation dataset $D_{vd}$. At the end of each dialogue session, CILK detects the set of bottom $\rho$\% query relations and entities in $\mathcal{P}$ based on MRR scores evaluated on the validation dataset. We call these sets the \textbf{diffident} relation and entity sets respectively \textit{for the next dialogue session}. If the query relation and/or entity issued in the next session belongs to the \textit{diffident} relation or entity set, CILK asks the user for supporting facts\footnote{Note, if (unknown) $e$ or $r$ appears the first time in a user query, then it cannot be in the diffident set. But the system has to ask the user question by default.}. Otherwise, it proceeds with inference, answering or rejecting the query. 

Algorithm 1 shows the interactive knowledge learning and inference process of CILK on a query $q_j=(e, r, ?)~or~(?, r, e)$ in a given dialogue session-$j$. Let $\mathcal{K}_j$, $\mathcal{P}_j$, $\mathcal{T}_j$ and $\mathcal{M}_j$ be the current version of KB, performance buffer, threshold buffer and inference model of CILK \textit{at the point when session-$j$ starts}. Then, the interactive knowledge learning and inference proceeds as follows:

\vspace{1mm}
$\bullet$~~If $r \notin \mathcal{K}_j$ or $r$ is diffident in $\mathcal{P}_j$, the interaction module $\mathcal{I}$ of CILK asks the user to provide clue(s) $C_r$ involving $r$ [Line 1-3]. Similarly, if $e \notin \mathcal{K}_j$ or $e$ is diffident in $\mathcal{P}_j$, $\mathcal{I}$ asks the user to provide entity fact(s) $F_e$ involving $e$ [Line 4-6]. 

$\bullet$~~If the user provides $C_r$ and/or $F_e$, $\mathcal{I}$ augments $\mathcal{K}_j$ with triples from $C_r$ and $F_e$ respectively and $\mathcal{K}_j$ expands to $\mathcal{K}_{j+1}$ [Line 7-12]. In this process, $\alpha$ \% of the triples in $C_r$ and $F_e$ are randomly marked as \textit{training} triples and rest $(1-\alpha)$\% are marked as \textit{validation} triples while storing them in $\mathcal{K}_j$. 

$\bullet$~~Next, a set of training triples $D_{tr}^r$, $D_{tr}^e$ and a set of validation triples $D_{vd}^r$, $D_{vd}^e$ are sampled randomly from $\mathcal{K}_{j+1}$ involving $r$ and $e$ respectively [Line 13-14] for training and evaluating $\mathcal{M}_j$. While sampling, we set the ratio of number of training triples to that of validation triples as $\alpha$ to maintain a fixed training and validation set distribution. The size for $(D_{tr}^r\cup D_{tr}^e)$ is set at most $N_{tr}$ (tuned based on real-time training requirements). 

$\bullet$~~Next, $\mathcal{M}_j$ is trained with $(D_{tr}^r \cup D_{tr}^e)$ and gets updated to $\mathcal{M}_{j+1}$ [Line 15]. Note that, training $\mathcal{M}_j$ with $(D_{tr}^r \cup D_{tr}^e)$ encourages $\mathcal{M}_j$ to learn the embeddings of both $r$ and $e$ before inferring $q_j$. Then, we evaluate $\mathcal{M}_{j+1}$ with $(D_{vd}^r ~\cup~ D_{vd}^e)$ in order to update the performance buffer $\mathcal{P}_j$ into $\mathcal{P}_{j+1}$ and threshold buffer $\mathcal{T}_j$ into $\mathcal{T}_{j+1}$ [Line 16]. Finally, $\mathcal{M}_{j+1}$ is invoked by CILK to either infer $q_j$ for predicting an answer entity $\widetilde{e}$ from $\mathcal{K}_{j+1}$ [Line 17] or reject $q_j$ to indicated that the true answer does not exist in $\mathcal{K}_{j+1}$. Note, CILK trains $\mathcal{M}_j$ and infers $q$ [Line 13-17] only if $e, q \in \mathcal{K}_{j+1}$.

\section{Experiments}

As indicated earlier, the proposed CILK system is best used in a \textit{multi-user} environment, so it naturally observes many more query triples (hence, accumulates more facts) from different users over time. Presently CILK fulfills its knowledge learning requirement by only adding the \textit{supporting facts} into the KB. The predicted query triples are not added as they are unverified knowledge. However, in practice, CILK can store these predicted triples in the KB as well after checking their correctness through \textit{cross-verification} while conversing with other users in some future related conversations by smartly asking them. 
Note that CILK may not verify its prediction with the same user who asked the question/query $q$ because he/she may not know the answer(s) for $q$. However, there is no problem that it acquires the correct answer(s) of $q$ when it asks $q$ to some other user $u'$ in a future related conversation and $u'$ answers $q$. At this point, CILK can incorporate $q$ into its KB and also, train itself using triple $q$. We do not address the issue here. 

\subsection{Evaluation Setup}
Evaluation of CILK with real users in a crowd-source based setup would be very difficult to conduct and prohibitively time-consuming (and expensive) as it needs a large number of real-time and continuous user interaction. Thus, we design a simulated interactive environment for the evaluation. 

\begin{table}[t]
	\scriptsize
	\centering
	\caption{\small Dataset statistics [\textit{kwn} = known, \textit{unk} = unknown]}
	\label{my-label}
	\begin{tabular}{|p{2.7cm}|c|c|c|c|c|c|}
		\hline
		\multicolumn{1}{|c|}{\textbf{KB Statistics}} & \multicolumn{3}{c|}{\textbf{WordNet}} & \multicolumn{3}{c|}{\textbf{Nell}} \\ \hline
		\# Relations ($\mathcal{K}_{org}$ / $\mathcal{K}_{b}$ )           & \multicolumn{3}{c|}{18 / 12}          & \multicolumn{3}{c|}{150 / 142}               \\ 
		\# Entities ($\mathcal{K}_{org}$ / $\mathcal{K}_{b}$)            & \multicolumn{3}{c|}{13, 595 / 13, 150} & \multicolumn{3}{c|}{11, 443 / 10, 547}           \\ 
		\# Triples ($\mathcal{K}_{org}$ / $\mathcal{K}_{b}$)             & \multicolumn{3}{c|}{53, 573 / 33, 159} & \multicolumn{3}{c|}{66, 529 / 51,252}        \\ \hline
		\# Test relations (\textit{kwn} / \textit{unk})           & \multicolumn{3}{c|}{18 (12 / 6)}           & \multicolumn{3}{c|}{25 (17 / 8)}           \\ 
		\# initial Train / intial valid / test (or query) triples ($D_q$) & \multicolumn{3}{c|}{29846~/~3323~/~1180} & \multicolumn{3}{c|}{46056 / 5196 / 1250}       \\ \hline
		\multicolumn{7}{|c|}{\textbf{Test (or query) triples ($D_q$) statistics [$(e, r, ?)$ or $(?, r, e)$]
		}} 
		\\ \hline
		\% triples with \textbf{only $e$ \textit{unk}}  & \multicolumn{3}{c|}{8.05}       & \multicolumn{3}{c|}{19.36}  \\
		\% triples with \textbf{only $r$ \textit{unk}} & \multicolumn{3}{c|}{30.25}       & \multicolumn{3}{c|}{21.84}  \\
		\%~triples \textbf{both $e$ and $r$ \textit{unk}} & \multicolumn{3}{c|}{5.25}       & \multicolumn{3}{c|}{10.16}  \\ \hline
	\end{tabular}
	\normalsize
\end{table}

We create a \textbf{simulated user} (a program) to interact with \textbf{CILK}, where the simulated user issues a query to CILK and CILK answers the query. The (simulated) user has (1) a \textbf{\textit{knowledge base}} ($\mathcal{K}_u$) for answering questions from CILK, and (2) an \textbf{\textit{query dataset}} ($D_q$) from which the user issues queries to CILK.\footnote{Using $\mathcal{K}_u$ and $D_q$, we can create \textbf{simulated dialogues} as well. Utterances in a dialogue can be created using a language template for each triple. Likewise, extraction of triples from utterances can be done using templates as well.} Here, $D_q$ consists of a set of structured query triples $q$ of the form ($e$, $r$, ?) and (?, $r$, $e$) readable by CILK. In practice, the user only issues queries to CILK, but cannot evaluate the performance of the system unless the user knows the answer. To evaluate the performance of CILK on $D_q$ in the simulated setting, we also collect the answer set for each query $q \in D_q$ (discussed shortly). 

As CILK is supposed to perform continuous online knowledge acquisition and learning, we evaluate its performance on the streaming query dataset. We assume that, CILK has been deployed with an initial knowledge base ($\mathcal{K}_b$) and the inference model $\mathcal{M}$ has been trained over all triples in $\mathcal{K}_b$ for a given number of epochs $N_{init}$. We call $\mathcal{K}_b$ the \textbf{base KB} of CILK which serves as its knowledge base at the time point ($t_{eval}$) when our evaluation starts. And the training process of $\mathcal{M}$ using triples in $\mathcal{K}_b$ is referred to as the \textbf{\textit{initial training phase}} of CILK onwards.~In the initial training phase, we randomly split $\mathcal{K}_b$ triples into a set of training triples $D_{tr}$ and a set of validation triples $D_{vd}$ with 9:1 ratio (we use $\alpha=0.9$) and train $\mathcal{M}$ with $D_{tr}$. $D_{vd}$ is used to tune model hyper-parameters and populate initial performance and threshold buffers $\mathcal{P}$ and $\mathcal{T}$ respectively. $D_{tr}$, $D_{vd}$, $\mathcal{P}$, and $\mathcal{T}$ get updated continuously after $t_{eval}$ in the \textbf{online training and evaluation phase} (with new acquired triples) during interaction with the simulated user.

\begin{table*}[t!]
\centering
\scriptsize
\caption{\small Comparison of predictive performance of various versions of CILK. For each KB dataset, we compare the first four (Threshold) variants denoted ase``X-BTr" and last three (dataset sampling strategy) variants denoted as ``MaxTh-X" and marked the highest H@1 and H@10 values (among each of the groups of four and three) in bold. Thus, some columns have at max. two values marked bold (due to the two comparison groups). \textbf{\textit{MaxTh-BTr} in the table is the version of CILK proposed in Sec.~3}.} 

\begin{tabular}{|l|ccc|ccc|ccc|ccc||ccc|}
\hline
\multirow{2}{*}{} & \multicolumn{3}{c|}{Rel - K / Ent -K}           & \multicolumn{3}{c|}{Rel - K /  Ent -UNK}        & \multicolumn{3}{c|}{Rel - UNK / Ent - K} & \multicolumn{3}{c||}{Rel - UNK / Ent -UNK} & \multicolumn{3}{c|}{Overall}                                                    \\ \cline{2-16} 
                  & MRR           & H@1            & H@10           & MRR           & H@1            & H@10           & MRR    & H@1            & H@10           & MRR    & H@1             & H@10           & \multicolumn{1}{l|}{MRR} & \multicolumn{1}{l|}{H@1} & \multicolumn{1}{l|}{H@10} \\ \hline

\multicolumn{16}{|c|}{\textbf{WordNet}}                                                                                                                                                                                                                                                                 \\ \hline
EntTh-BTr           & 0.46          & 34.57          & 57.23          & 0.04        & 3.50           & 4.38           & 0.20   & \textbf{16.21}  &   25.80          & 0.07   & 4.83            & 8.06           & 0.33                     & 25.03                    & 40.89                     \\ 
RelTh-BTr           & 0.45          & 12.71 &  16.32          & 0.04          & \textbf{7.89}           & \textbf{7.89}           & 0.21   & 12.30          & 16.51          & 0.07   & \textbf{9.67}   & \textbf{9.67}  & 0.33           & 12.09           & 15.39                     \\ 
MinTh-BTr           & 0.45          & 33.81          & \textbf{57.99}  & 0.03          & 2.63           & 3.50           & 0.22  & 15.93          & \textbf{28.05}          & 0.07  & 4.84             & 8.06           & 0.33                     & 24.43                    & \textbf{41.91}                     \\ \hline 
MaxTh-BTr           & 0.45          & \textbf{34.72} & \textbf{56.87}  & 0.04      & 5.26  & 6.14  & 0.20   & \textbf{15.92} & \textbf{25.79} & 0.07   & 6.45            & \textbf{9.67}           & 0.33                     & \textbf{25.27}           & \textbf{40.95}            \\ \hline
MaxTh-EntTr           & 0.42         & 26.07          & 42.74          & 0.26          & \textbf{19.29}          & \textbf{22.80}          & 0.19  & 11.79          & 15.17          & 0.23   & \textbf{17.74}            & \textbf{20.96}           & 0.33                     & 20.77                    & 31.60                     \\ 
MaxTh-RelTr           & 0.45         & 34.48          & 55.93          & 0.003          & 2.63           & 3.51           & 0.13  & 11.25          & 18.01          & 0.11   & 8.06  & 16.13 & 0.30                     & 23.46                    & 38.09                     \\ \hline

\multicolumn{16}{|c|}{\textbf{Nell}}                                                                                                                                                                                                                                                                    \\ \hline
EntTh-BTr           & 0.37          & 26.80          & 47.28          & 0.06         & 4.47            & 7.22            & 0.15   & 9.58           & 19.97          & 0.04   & 1.64            & 7.36           & 0.22                     & 16.18                    & 29.78                     \\ 
RelTh-BTr           & 0.37          & 17.01          & 25.05          & 0.06 & 3.78           & 4.13           & 0.16   & 8.72           & 17.67          & 0.03   & 3.28   & 4.92           & 0.23                     & 11.35                    & 17.49                     \\ 
MinTh-BTr           & 0.37          & 26.63          & 47.30          & 0.06          & \textbf{5.33}           & \textbf{8.60}           & 0.15   & 10.24 & \textbf{23.21} & 0.03   & 1.64            & 5.72           & 0.23                     & 16.41                   & \textbf{30.57}            \\ \hline

MaxTh-BTr         & 0.37  & \textbf{27.57}  & \textbf{47.58} & 0.06 & \textbf{4.30}  & 7.57   & 0.16   & \textbf{10.69}           & 19.61          & 0.03   & \textbf{4.92}            & \textbf{8.20}  & 0.23                     & \textbf{17.16}           & \textbf{30.03}            \\ \hline
MaxTh-EntTr           & 0.34          & 21.82          & 42.65          & 0.13          & 3.95           & \textbf{7.91}           & 0.22   & \textbf{16.48}           & \textbf{20.56}          & 0.06   & 4.06   & 4.06  & 0.24                     & 15.46                    & 27.44                     \\ 
MaxTh-RelTr           & 0.37          & 26.60          & 47.07          & 0.04          & 3.44           & 5.85           & 0.20   & 12.18 & 17.67   & 0.06   & 3.28            & \textbf{10.67}          & 0.23                     & 16.67                    & 29.29                     \\ \hline
\end{tabular}
\vspace{-0.3cm}
\end{table*}

The relations and entities in $\mathcal{K}_b$ are regarded as \textit{known} relations and \textit{known} entities to CILK till $t_{eval}$. Thus, the initial inference model $\mathcal{M}$ is trained and validated with triples involving only \textit{known} relations and \textit{known} entities (in $\mathcal{K}_b$). During the online training and evaluation phase, CILK faces queries (from $D_q$) involving both \textit{known} and \textit{unknown} relations and entities. More specifically, if a relation (entity) appearing in a query $q \in D_q$ exists in $\mathcal{K}_b$, we consider that query relation (entity) as \textit{known} query relation (entity). Otherwise, it is referred to as \textit{unknown} query relation (entity).

We create simulated user's KB $\mathcal{K}_u$, base KB ($\mathcal{K}_b$) and query dataset $D_q$ from two standard KB datasets: (1) \textbf{WordNet} \cite{bordes2013translating} and (2) \textbf{Nell} \cite{gardner2014incorporating}. From each KB dataset, we first build a fairly large triple store and use it as the original KB ($\mathcal{K}_{org}$) and then, create $\mathcal{K}_u$ of user, base KB ($\mathcal{K}_b$) of CILK and $D_q$ from $\mathcal{K}_{org}$, as discussed below (Table 1 shows the results).

\vspace{1mm}
\noindent
\textbf{Simulated User, Base KB Creation and Query Dataset Generation.}  
In Nell, we found 150 relations with $\geq 300$ triples, and we randomly selected 25 relations for $D_q$. We shuffle the list of 25 relations, select 34\% of them as \textit{unknown} relations and consider the rest (66\%) as \textit{known} relations. 

For each \textit{known} relation $r$, we randomly shuffle the list of distinct triples for $r$, choose (maximum) 250 triples and randomly select 20\% as test and add a randomly chosen subset of the rest of the triples along with the leftovers (not in the list of 250), into $\mathcal{K}_{b}$ and the other subset are added to $\mathcal{K}_{u}$ (to provide supporting facts involving poorly learned \textit{known} relations and/or \textit{entities}, if asked [see Sec 3.3]).  

For each \textit{unknown} relation $r$, we remove all triples of $r$ from $\mathcal{K}_{org}$, randomly choose 20\% triples among them and reserve them as query triples for unknown $r$. Rest 80\% triples of \textit{unknown} $r$ are added to $\mathcal{K}_{u}$ (for providing clues). In this process, we also make sure that the query instances involving \textit{unknown} $r$ are excluded from $\mathcal{K}_{u}$. Thus, the user cannot provide the query triple itself as a clue to CILK (during inference) and also, to simulate the case that the user does not know the answer of its issued query. 
Note, if the user cannot provide a clue for an unknown query relation or a fact for an unknown query entity (not likely), CILK will not be able to correctly answer the query. 

\begin{table}[t!]
\centering
\scriptsize
\caption{\small Performance of CILK Threshold variants on Rejection and prediction decisions. Here, AE ($\neg$AE) means true answer entity exists (does not exist) in KB. ``Pr(pred$\mid$AE)" means the probability of predicting an answer, given the true answer exists in KB. ``Pr(Reject$\mid\neg$AE)" means probability of rejecting the query, given true answer does not exist in KB.}
\begin{tabular}{|l|c|c|c|c|}
\hline
\multirow{2}{*}{}               & \multicolumn{2}{c|}{WordNet}                                                                                                                               & \multicolumn{2}{c|}{Nell}                                                                                                                                   \\ \cline{2-5} 
                                & Pr(pred$\mid$AE) & \begin{tabular}[c]{@{}c@{}}Pr(Reject $\mid$ \\ $\neg$AE)\end{tabular} & Pr(pred$\mid$AE) & \begin{tabular}[c]{@{}c@{}}Pr(Reject $\mid$ \\ $\neg$AE)\end{tabular} \\ \hline
EntTh-BTr                       & 0.85                                                            & 0.24                                                                                     & 0.82                                                            & 0.15                                                                                      \\ 
RelTh-BTr                       & 0.20                                                            & \textbf{0.92}                                                                            & 0.26                                                            & \textbf{0.72}                                                                             \\ 
\multicolumn{1}{|c|}{MinTh-BTr} & \textbf{0.90}                                                   & 0.18                                                                                     & \textbf{0.86}                                                   & 0.10                                                                                      \\ 
MaxTh-BTr                       & 0.83                                                            & 0.33                                                                                     & 0.72                                                            & 0.31                                                                                     \\ \hline
\end{tabular}
\vspace{-0.15cm}
\end{table}

At this point, $D_q$ consists of query triples involving both \textit{known} and \textit{unknown} relations, but all \textit{known} entities. To create queries in $D_q$ having unknown entities, we randomly choose 20\% of the entities 
in $D_q$ triples, remove all triples involving those entities from $\mathcal{K}_{org}$ and add them to $\mathcal{K}_{u}$. Now, $\mathcal{K}_{org}$ gets reduced to $\mathcal{K}_b$ (base KB). Next, for each query triple ($h$, $r$, $t$) $\in D_q$, we convert the triple into a head query $q =$(?, $r$, $t$) [or a tail query $q =$($h$, $r$, ?)] by randomly deleting the head or tail entity. We also collect the answer set for each $q \in D_q$ based on observed triples in $\mathcal{K}_{org}$ for CILK evaluation. Note, the generated query triples (with answer entity) in $D_q$ are not directly in $\mathcal{K}_b$ or $\mathcal{K}_{u}$.

The WordNet dataset being small, we use all its 18 relations for creating $D_q$, $\mathcal{K}_{u}$, $\mathcal{K}_{b}$ following Nell. As mentioned earlier, the triples in $\mathcal{K}_{b}$ are randomly split into 90\% training and 10\% validation datasets for simulating \textit{initial training phase} of CILK.

\vspace{0.5mm}
\noindent
\textbf{Hyper-parameter Settings.} 
Embedding dimensions of entity and relations are empirically set as 250 for WordNet and Nell, initial training epochs $N_{init}$ as 100 for WordNet (140 for Nell), training batch size 128, $N_{tr}$ as 500,  $|D_{vd}^r \cup D_{vd}^e|$ as 50, $\alpha=0.9$, $\rho=20\%$, random seed as 1000, 4 negative triples generated per positive triple, online training epoch as 5 (2) for each closed (open) world query processing, and learning rate 0.001 for both KB datasets. L2-regularization parameter set as 0.001. Adam optimizer is used for optimization.

\vspace{0.5mm}
\noindent
\textbf{Compared Models.} Since there is no existing work that solves our proposed problem, we compare various versions of CILK, constructed based on different types of prediction threshold $\mu_q$ for query rejection (Sec. 3.2) and various online training $D_{tr}=(D_{tr}^r \cup D_{tr}^e)$ and validation dataset $D_{vad}=(D_{vd}^r \cup D_{vd}^e$) sampling strategies [see Line 13-14 of Algorithm 1] as discussed below: 

\vspace{1mm}
$\bullet$~~\textbf{CILK variants based on prediction threshold types,} namely \textit{EntTh-BTr}, \textit{RelTh-BTr}, \textit{MinTh-BTr} and \textit{MaxTh-BTr}~ (see Table 2). For \textit{EntTh-BTr}, we use $\mu_q = max\{\mathcal{T}[e],0\}$, for \textit{RelTh-BTr}, we use $\mu_q = max\{\mathcal{T}[r], 0\}$, for \textit{MinTh-BTr}, we use $\mu_q = max\{min\{\mathcal{T}[e], \mathcal{T}[r]\}, 0\}$ and \textit{MaxTh-BTr} uses $\mu_q =max\{\mathcal{T}[e], \mathcal{T}[r], 0\}$ as proposed in Sec 3.2. Here, ``\textit{BTr}" indicates that the CILK variant samples triples involving both query entity and relation from KB to build $D_{tr}$ and $D_{vd}$.

$\bullet$~\textbf{CILK variants based on dataset sampling strategies:} \textit{MaxTh-BTr} (as explained above), \textit{MaxTh-EntTr} and \textit{MaxTh-RelTr} (see Table 2). Given the query entity $e$ and query relation $r$, \textit{MaxTh-EntTr} only samples triples involving $e$ and \textit{MaxTh-RelTr} samples only triples involving $r$ to build $D_{tr}$ and $D_{vd}$. Note, if the sampled dataset $D_{tr}$ ($D_{vd}$) is $\emptyset$, CILK skips online training (validation) steps for that session.

\begin{table}[t!]
\centering
\scriptsize
\caption{\small Overall Performance of \textbf{MaxTh-BTr} (CILK), varying the maximum number of clues (\#C) and entity facts (\#EF) acquired from user per dialogue session (if asked by the interaction module $\mathcal{I}$).
}
\begin{tabular}{|l|ccc||ccc|}
\hline
\multirow{2}{*}{\begin{tabular}[c]{@{}l@{}}(\#C, \\ \#EF)\end{tabular}} & \multicolumn{3}{c||}{WordNet}                    & \multicolumn{3}{c|}{Nell}                       \\ \cline{2-7} 
                                                                        & MRR           & H@1            & H@10           & MRR           & H@1            & H@10           \\ \hline
(1, 1)         & 0.30          & 22.09          & 37.83          & 0.23          & 16.89          & 31.14          \\ 
(1, 2)     & 0.32          & 23.00          & 39.25          & \textbf{0.25}         & \textbf{18.11}          & \textbf{31.30}          \\ \hline 
(1, 3)                                                                  & \textbf{0.33}          & \textbf{25.27} & \textbf{40.95} & 0.23          &    17.16   &  30.03 \\ \hline
(1, 3)-U           & 0.31                               & 23.52                               & 38.15                               & 0.21                      & 15.77                               & 28.64                \\              

(2, 2)       & 0.32 & 23.43 & 39.05          & 0.23 &  16.82 & 30.33 \\ \hline
\end{tabular}
\vspace{-0.15cm}
\end{table}

\vspace{1mm}
\noindent
\textbf{Evaluation Metrics.}~We use two common KBE evaluation metrics: \textit{mean reciprocal rank} (MRR)  and \textit{Hits@\textit{k}} (H@\textit{k}). MRR is the average inverse rank of the top ranked true answer entity for all queries \cite{bordes2013translating}. Hits@\textit{k} is the proportion of test queries for which the true answer entity has appeared in top-\textit{k} (ranked) predictions. Higher MRR and Hits@\textit{k} indicate better performance.

\vspace{-0.1cm}
\subsection{Results and Analysis}
For evaluation on a given KB (WordNet or Nell), we randomly generate a chronological ordering of all query instances in $D_q$, which are fed to the trained CILK (after the initial training phase is over) in a streaming fashion, and then evaluate CILK on the overall query dataset. The avg. test query processing time of CILK is 1.25 sec (on a Nvidia Titan RTX GPU). While evaluating a query $q_j$, if the true answer of $q_j$ does not exist in KB $\mathcal{K}_{j+1}$ and $\mathcal{M}_{j+1}$ rejects $q_j$, we consider it as a correct prediction. For such $q_j$, Reciprocal Rank (RR) cannot be computed. Thus, we exclude $q_j$ while computing MRR, but consider it in computing Hits.

\begin{table}[t!]
\centering
\scriptsize
\caption{\small Performance of \textbf{MaxTh-BTr} (CILK) on test queries observed over time, given the model has made a prediction.}
\begin{tabular}{|c|c|c|c|c|c|c|}
\hline
\multirow{2}{*}{\begin{tabular}[c]{@{}c@{}}\% Test Data\\ Observed\end{tabular}} & \multicolumn{3}{c|}{WordNet}                               & \multicolumn{3}{c|}{Nell}                                   \\ \cline{2-7} 
                                                                                    & MRR           & H@1            & \multicolumn{1}{l|}{H@10} & MRR            & H@1            & \multicolumn{1}{l|}{H@10} \\ \hline
\multicolumn{7}{|c|}{Overall Performance}                                                                                                                                                                      \\ \hline
50\%         &    \textbf{0.37}   &  27.50   &  \textbf{47.19}             & \textbf{0.29}   & 20.77 & \textbf{38.87}            \\ 
100\%        &  \textbf{0.37}    & \textbf{27.67}         & 46.71            & \textbf{0.29}           & \textbf{20.82}          & 38.65                     \\ \hline
\multicolumn{7}{|c|}{On Open-word Queries}                                                                                                                                                                     \\ \hline
50\%      & 0.16          & 11.87           & 20.11         & 0.09          & 4.81           & 16.47                     \\ 
100\%     & \textbf{0.18} & \textbf{12.90}  & \textbf{22.91}   & \textbf{0.13} & \textbf{8.58}  & \textbf{19.54}            \\ \hline
\end{tabular}
\vspace{-0.15cm}
\end{table}

Table 2 shows the performance of CILK variants on the query dataset, evaluated in terms of MRR, H@1 and H@10 for both KBs. We present the overall result on the whole query dataset as well as results on subsets of query datasets, denoted as (\textit{Rel-X, Ent-Y}), where X and Y can be either \textit{known} (`K') or \textit{unknown} (`UNK') and `\textit{Rel}' denotes query relation and `\textit{Ent}' denotes query entity. So, here, (\textit{Rel-K, Ent-UNK}) denotes the subset of the query dataset that contains query triples involving only known query relations and unknown query entities (with respect to $\mathcal{K}_b$). For all variants, we fix the maximum number of clue triples and entity fact triples provided by the simulated user for each query (when asked) as 1 and 3 respectively.

From Table 2, we see that, \textit{MaxTh-BTr} (version of CILK in Sec.~3) achieves the overall best results compared to other variants for both KB datasets. Among different threshold versions, \textit{MaxTh-BTr} and \textit{MinTh-BTr} perform better than the rest. The relatively poor result of \textit{RelTh-BTr} shows threshold strategy plays a vital role in performance improvement. Considering different dataset sampling strategies, again we see \textit{MaxTh-BTr} performs better than other versions. As the triples involving both query entity and relation are selected for online training in \textit{MaxTh-BTr}, CILK gets specifically trained on relevant (query-specific) triples before the query is answered. For other variants, either triples involving query relation (for \textit{MaxTh-EntTr}) or triples involving query entity (for \textit{MaxTh-RelTr}) are discarded, causing a drop in performance.

In Table 3, we compare different CILK threshold variants based on how often it predicts (or rejects) the query, when the true answer exists (does not exist) in its current KB, given by Pr(pred $\mid$ AE) [ Pr(Reject $\mid$ $\neg$AE) ]. For both datasets, \textit{EntTh-BTr} has a tendency to predict more and reject less. Whereas, \textit{RelTh-BTr} is more precautious in prediction. \textit{MinTh-BTr} is the least precautious in prediction among all. \textit{MaxTh-BTr} adopts the best of both worlds (\textit{EntTh-BTr} and \textit{RelTh-BTr}), showing moderate strategy in prediction and rejection behavior.

Table 4 shows comparative performances of \textit{MaxTh-BTr} on varying the maximum number of clue triples and entity fact triples provided by the user (when asked). Comparing (1, 1), (1, 2), (1, 3) we see a clear performance improvement in \textit{MaxTh-BTr} with the increase in (acquired) entity fact triples (specially, for WordNet). This shows that if user interacts more and provides more information for a given query, CILK can gradually improve its performance over time [i.e., with more accumulated triples in its KB]. For Nell, performance improves for both (1, 2) and (1, 3) compared to that in (1, 1), (1, 2) variant being the best overall. Comparing (1, 3) and (2, 2) for both KBs, we see that acquiring more entity facts dominates the overall performance improvement compared to acquiring more clues. This is because, as a past query relation is more probable to appear in future query compared to a past query entity, CILK can gradually learn the relation embedding with less clues per query unlike that for an entity. (1, 3)-U denotes the set up, where CILK asks for clues or entity facts only if the query triple has unknown entity and/or relation, i.e. we disable the use of performance buffer $\mathcal{P}$ (see Sec 3.3). Due to lack of sufficient training triples to learn an unknown query relation and entity, the overall performance degrades. This shows the importance and effectiveness of the performance buffer in improving performance of CILK with limited user interactions.

In Table 5, we show the performance of MaxTh-BTr on (predicted) test queries over time. Considering overall performance, the improvement is marginal. However, for open-world queries, there is a substantial improvement in performance as CILK relatively acquires more facts for open-world queries than that of closed-world ones.

\vspace{-0.1cm}
\section{CILK: Use Cases in Dialogue Systems}
\vspace{-0.1cm}
There are many applications for CILK.  Conversational QA systems \cite{kiyota2002dialog,bordes2014question}, conversational recommendation systems \cite{anelli2018knowledge,zhang2018towards}, 
information-seeking conversational agents \cite{yang2018response}, etc., that deal with real-world facts, are all potential use cases for CILK. 

Recently, \cite{Young2018Augmenting,zhou2018commonsense} showed that dialogue models augmented with commonsense facts improve dialogue generation performance. It's quite apparent that continuous knowledge learning using CILK can help these models grow their KBs over time and thereby, improve their response generation quality.

The proposed version of CILK has been designed based on a set of assumptions (see Sec. 1) to reduce the complexity of the modeling. For example, we do not handle the case of intentional or unintentional false knowledge injection by users to corrupt the system's KB. Also, we do not deal with fact extraction errors of the peripheral information extraction module or query parsing errors of the semantic parsing modules, which can affect the knowledge learning of CILK. We believe these are separate research problems and are out of the scope of this work. In future, we plan to model an end-to-end approach of knowledge learning where all peripheral components of CILK can be jointly learned with CILK itself. We also plan to solve the cold start problem when there is little training data for a new relation when it is first added to the KB.  

Clearly, CILK does not learn all forms of knowledge. For example, it does not learn new concepts and topics, user traits and personality, and speaking styles. They also form a part of our future work. 


\section{Conclusion}
In this paper, we proposed a continuous (or lifelong) and interactive knowledge learning engine CILK for dialogue systems. It exploits the situation when the system is unable to answer a WH-question from the user (considering its existing KB) by asking the user for some knowledge and based on it to infer the query answer. We evaluated the engine on two real-world factual KB data sets and observed promising results. This also shows the potentiality of CILK to serve as a factual knowledge learning engine for future conversational agents.

\section*{Acknowledgments}
This work was partially supported by a grant from
National Science Foundation (NSF IIS 1838770)
and a research gift from Northrop Grumman. 

\bibliography{acl2019}
\bibliographystyle{acl_natbib}

\end{document}